\documentclass[letterpaper, 10 pt, conference]{ieeeconf}  

\IEEEoverridecommandlockouts                              

\usepackage{graphicx} 

\usepackage{color} 
\usepackage{soul} 

\title{\LARGE \bf
A Robot Simulation Environment for Virtual Reality Enhanced Underwater Manipulation and Seabed Intervention Tasks*
}

\author{Sumey El-Müftü$^{1}$ and Berke Gür$^{2}$
\thanks{*This work was supported in part by The Scientific and Technological Research Council of Türkiye (TÜBİTAK) with the project title ``Robotic Squid for Underwater Manipulation and Intervention'' (grant no. 216M201).}
\thanks{$^{1}$Sumey El-Müftü was with the Mechatronics Engineering program of the Graduate School, Bahçeşehir University, Beşiktaş-İstanbul, Türkiye
        {\tt\small sumey.muftu@outlook.com}}%
\thanks{$^{2}$Berke Gür is with the Faculty of Engineering, Department of Mechanical Engineering, Marmara University, Maltepe-İstanbul, Türkiye
        {\tt\small berke.gur@marmara.edu.tr}}%
}

\begin{document}

\maketitle
\thispagestyle{empty}
\pagestyle{empty}

\begin{abstract}

This paper presents the (MARUN)$^2$ underwater robotic simulator. The simulator architecture enables seamless integration with the ROS-based mission software and web-based user interface of URSULA, a squid inspired biomimetic robot designed for dexterous underwater manipulation and seabed intervention tasks. (MARUN)$^2$ utilizes the Unity game engine for physics-based rigid body dynamic simulation and underwater environment modeling. Utilizing Unity as the simulation environment enables the integration of virtual reality and haptic feedback capabilities for a more immersive and realistic experience for improved operator dexterity and experience. The utility of the simulator and improved dexterity provided by the VR module is validated through user experiments.   

\end{abstract}

\section{INTRODUCTION} \label{Sec: intro}

Advancements in underwater robotic manipulation have paved the way for remote teleoperation and intervention in challenging aquatic environments. Several well-publicized recent developments have emphasized the increasing importance of dexterous underwater manipulation and intervention capabilities, in particular, for vehicles operating close to the seabed. In line with these developments, novel underwater robots specifically designed for such tasks have emerged over the recent years \cite{Rib12}-\cite{Man18}, including project URSULA. \cite{Gur24} 

The early underwater robotic simulators were extensions of general-purpose robot simulators such as the Gazebo \cite{gazebo} simulator of Robot Operating System (ROS). \cite{ros} The Unmanned Underwater Vehicle Simulator (UUVSim) \cite{Man16} and its newer derivative DAVE \cite{Zha22} are two such examples. More recently, game engines are being widely adopted for developing, testing, and optimizing underwater robot systems prior to real-world deployment. Potokar et al. \cite{Pot22}, \cite{Pot24} presented an underwater robot simulator based on the Unreal Engine (UE) \cite{unreal} game development environment. Although the simulator supports multiple sensors, emphasis was placed on simulating imaging sonar systems. A more comprehensive simulator, also based on UE software is described by Zhang et al. \cite{Zha23} In contrast, several simulators utilize the Unity simulation environment \cite{unity} as an alternative to UE. A complete Unity-based open source underwater robotics simulator that utilizes ROS as a middleware was presented by Chaudhary et al. \cite{Cha21} Focused on simulating on board sonar systems, the primary objective of this simulator was to provide a realistic virtual environment for the simulation of real-world remotely operated vehicle (ROV) operations, such as pipeline inspection. Lončar et al. \cite{Lon22} developed another Unity-based simulator that also uses ROS with an emphasis on synthetic data generation for machine learning applications. Collins et al. \cite{Col21} presented a concise review of some important marine robot simulators.

Typical underwater robotic manipulation systems on ROVs often rely on conventional 2-D screen interfaces, which limit operator depth perception, spatial awareness, and dexterity. Early research explored enhanced vision systems, including high-sensitivity cameras \cite{Ots92}, laser-based 3-D imaging \cite{Zhe92}, and line-scan imaging \cite{Gor93}, but these remained dependent on lighting and water turbidity. To address this, the researchers introduced synthetic background overlays and augmented visual interfaces \cite{Ode91}, combining video feeds with CAD-generated 3D models to improve navigation and spatial awareness, although misalignment challenges persist. More recently, virtual reality (VR) has been applied for immersive teleoperation \cite{Hal94}, allowing operators to interact with synthetic 3D underwater environments, enhancing navigation, depth perception, and collision avoidance. VR solutions have also been explored for mission planning and operator training \cite{Lan94, Hug90}. Our research group previously investigated touchless user interfaces (UI) to improve ROV operations \cite{Kap21}. The integration of VR into underwater teleoperation presents a promising alternative, providing an intuitive interface for robotic manipulation. A VR-enhanced simulation environment developed for project URSULA is based on the Unity game engine and ROS, which enhances dexterous underwater teleoperation. Recent advances in VR-based underwater teleoperation show improvements in user engagement, task efficiency, and ease of control over traditional 2-D interfaces, although challenges such as dizziness and adaptation time remain, necessitating further exploration in realistic simulation environments.

This paper presents the (MARUN)$^2$ robotic simulation environment, developed for project URSULA, to simulate VR and haptic feedback-enhanced dexterous underwater manipulation and seabed intervention. The paper is structured as follows. Section \ref{Sec: ursula} introduces project URSULA, followed by Section \ref{Sec: marun2}, which details the (MARUN)$^2$ simulator, including its integration with Unity, ROS, and VR capabilities. Section \ref{Sec: results} outlines the current applications of the simulator, while Section \ref{Sec: conclusions} concludes the paper with a discussion on planned future developments.

\section{PROJECT URSULA} \label{Sec: ursula}

\begin{figure}[thpb]
\centering
\includegraphics[width = 3.4in]{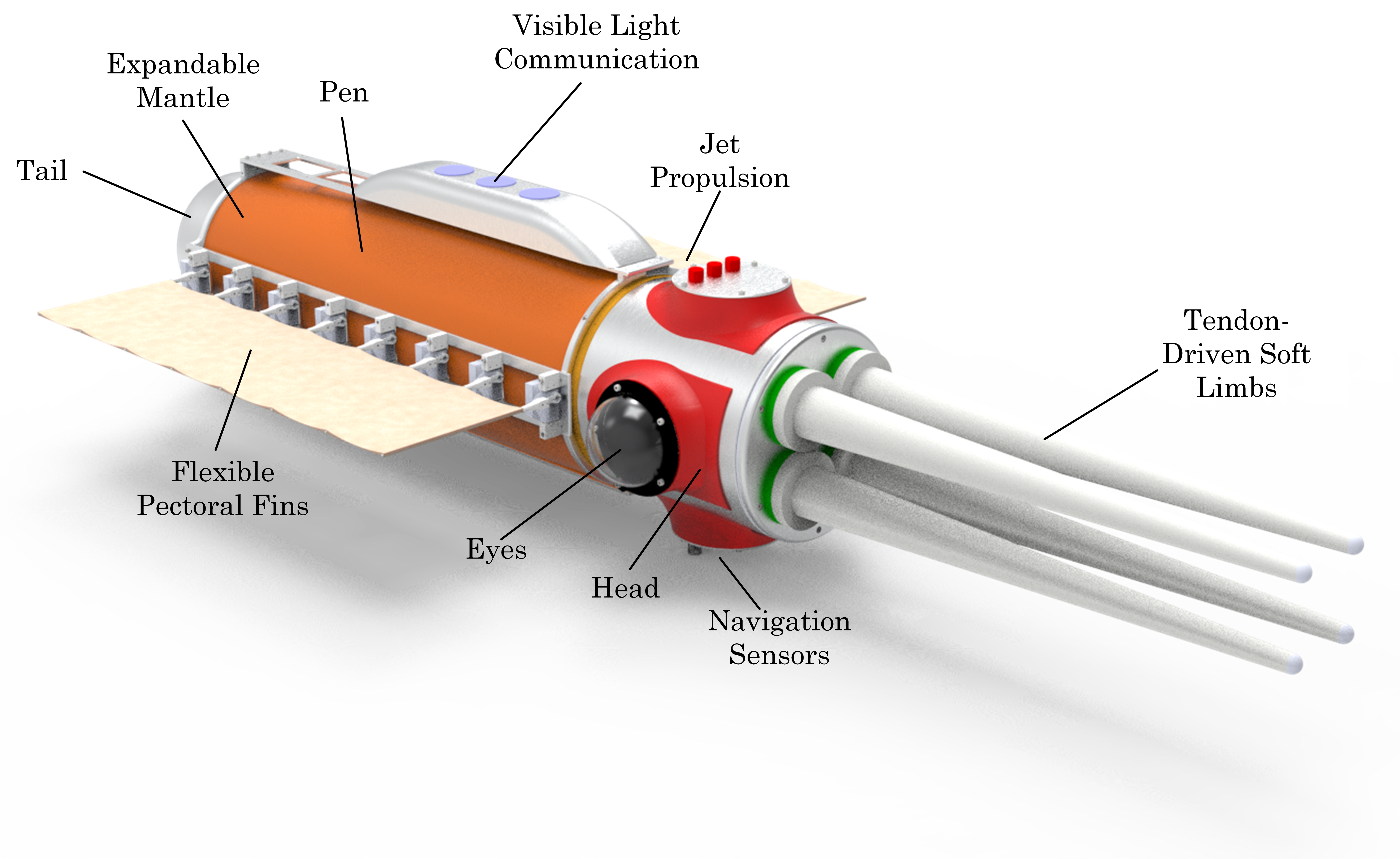}
\caption{A rendering of the production-ready 3-D solid model of URSULA highlighting key design aspects}
\label{fig: ursulaModel}
\end{figure}

In this section, the squid-inspired biomimetic robot platform designed for dexterous underwater manipulation and seabed intervention (dubbed {\textbf U}nmanned {\textbf R}obotic {\textbf S}quid for {\textbf U}nderwater and {\textbf L}ittoral {\textbf A}pplications, or project URSULA for short) is introduced. Inspired by the common squid and cuttlefish, URSULA is made up of 7 main body parts: 1) soft limbs, 2) head, 3) pen, 4) expandable mantle, 5) tail, 6) visible light communication system, and 7) flexible pectoral fins (see Fig. \ref{fig: ursulaModel}). The total length of the system (excluding the limbs) is less than 1200 mm, while the limbs are 600 mm long. The largest diameter is 250 mm at the head. The total dry weight of the URSULA is around 30 kg, and the robot is designed to be slightly positively buoyant. The head, pen, and tail form the rigid and watertight components of the main body. Tendon driven soft limbs are attached to the head. The head houses the tendon actuators, as well as several navigation sensors. The pen bridges the head to the tail, functioning as an internal shell that hosts the electronics housing and batteries. The connectors for the (optional) umbilical cable or tether and other waterproof connections are located at the tail. The port and starboard pectoral fins (each grouped into three sections) form the main actuation system of the robot and are attached to the two rails on the port and starboard sides that run along the length of the pen, outside of the mantle. These lateral rails are complemented by a pair of identical dorsal and ventral rails, which are used to attach several navigational sensors and the visible-light communication system. The visible light communication system, in the form of a dorsal fin, provides a wireless communication interface required for the haptic teleoperation of the limbs. The jet propulsion system is designated as the secondary propulsion system and is located towards the front part of the pen, inside of the expandable mantle. Seawater is pumped into the double-walled soft mantle. This causes the mantle to expand inward, increasing water pressure between the mantle and the pen. When sufficient water pressure is obtained, a solenoid valve is opened, forcing the water out through a steerable nozzle. 

\begin{figure}[thpb]
\centering
\includegraphics[width = 3.4in]{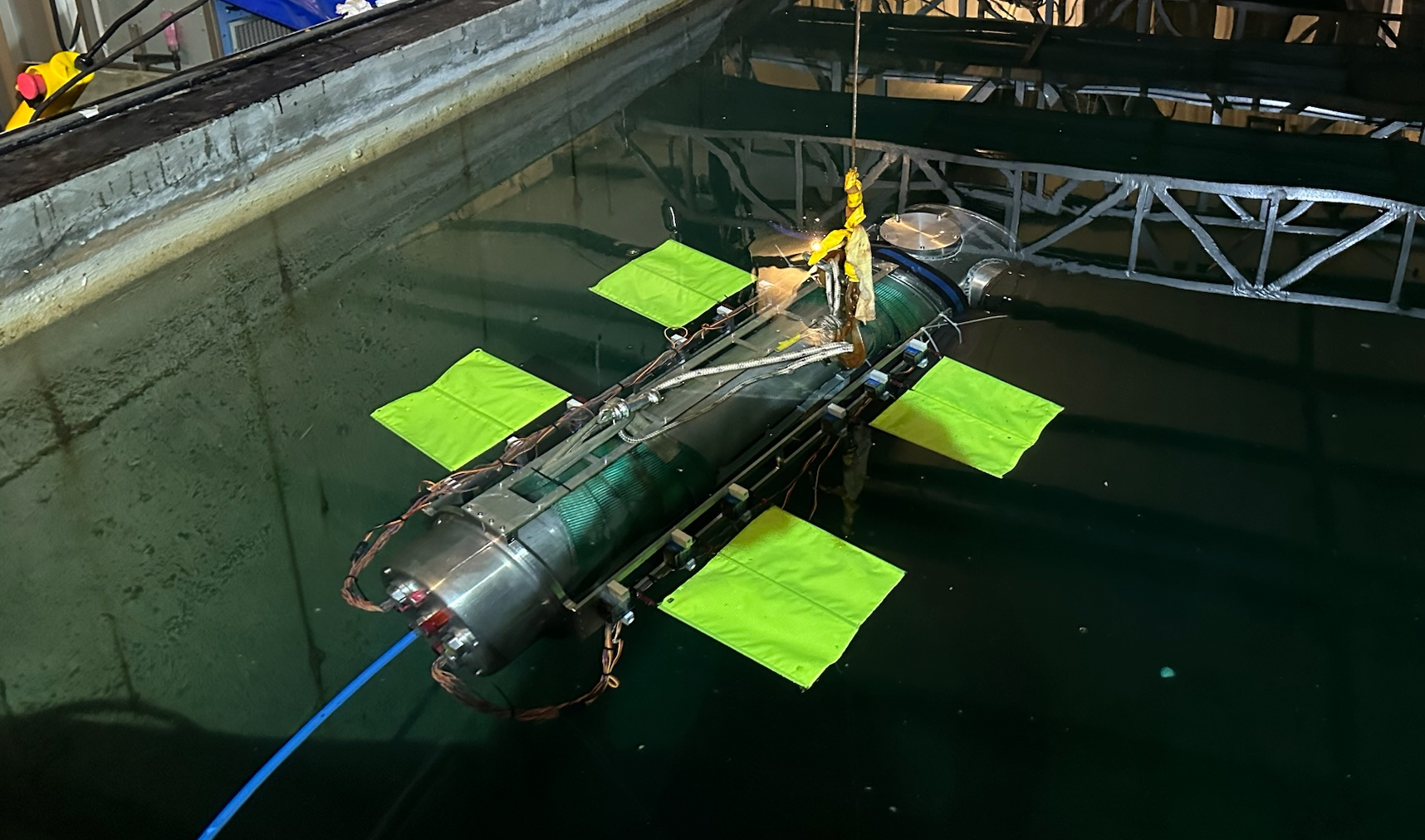}
\caption{The first prototype of URSULA undergoing preliminary pool tests}
\label{fig: ursulaPrototype}
\end{figure}

The primary motivation behind URSULA was to develop a robot that can dexterously interact with the underwater environment. Therefore, soft robotic limbs are the most crucial subsystem of the robot. In the current design, the robot has four soft, tendon-actuated robotic limbs protruding from the head. Two of these limbs are designated as arms and are designed to provide functionality such as reaching,
gripping, and pulling to perform manipulation tasks such as palpation, grasping, and retrieval. Force/pressure sensors are embedded into the tip of the arms for haptic force feedback. The remaining two limbs are designated as tentacles. These tentacles host an underwater camera and lighting system embedded at their distal ends. The camera system complements the on-board 3-D vision system for visual feedback-assisted teleoperation of the limbs as well as environment mapping. The motion of the light tentacle is synchronized with the camera tentacle to effectively illuminate the workspace of the arms. The tentacles are fully functional limbs and can also be used for other manipulation tasks such as anchoring the robot. Due to the highly elastic, soft material used in their construction, the robotic limbs can safely interact with the fragile environment in tasks such as palping and grasping, and handle delicate payloads and underwater objects. The first prototype of URSULA that underwent pool tests is depicted in Fig. \ref{fig: ursulaPrototype}.

\section{THE (MARUN)$^2$ SIMULATOR} \label{Sec: marun2}

\begin{figure}[thpb]
    \centering
    \includegraphics[width=3.4in]{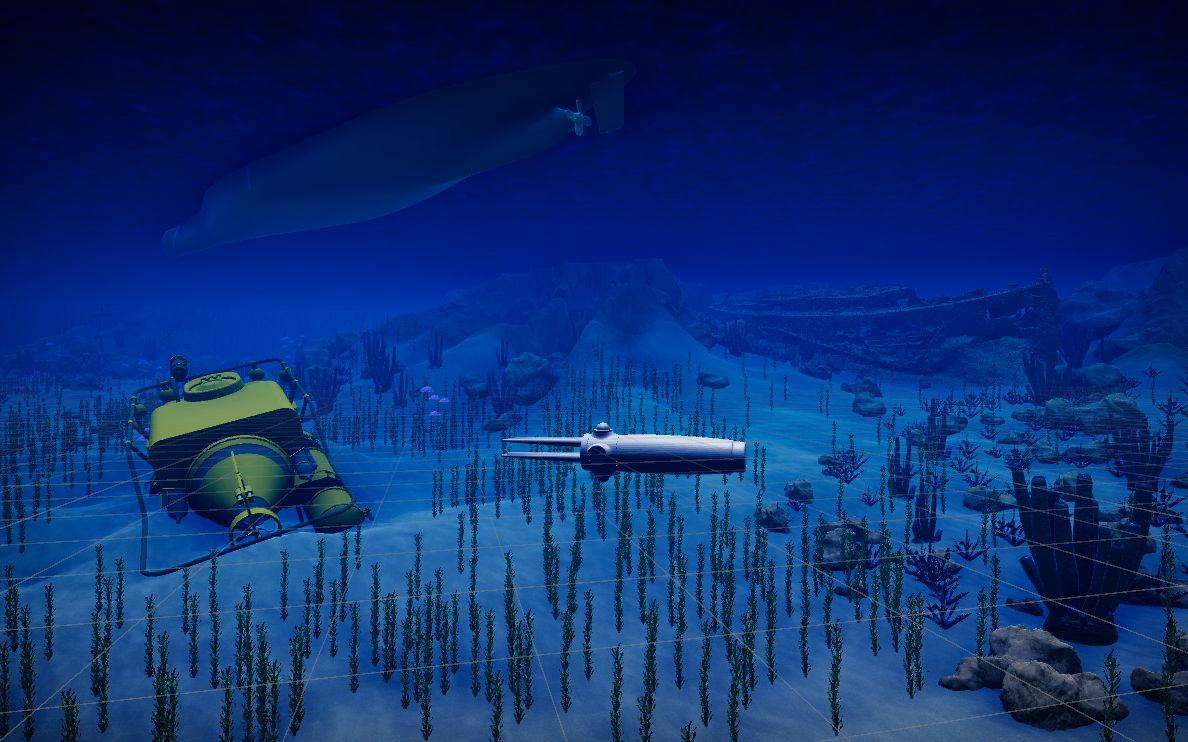} 
    \caption{Underwater environment depicting a sunken submarine, a surface vessel, and the  URSULA robot.}
    \label{fig: wreck}
\end{figure}

Designed as part of project URSULA, (MARUN)$^2$ (\textbf{Mar}mara \textbf{Un}iversity \textbf{Ma}rine \textbf{R}obotics \textbf{Un}ity) is a simulation environment for dexterous underwater manipulation and seabed intervention. (MARUN)$^2$ integrates VR, intuitive teleoperation, haptic feedback (planned for future versions) and a dynamic simulation environment to enhance operator control and seabed intervention capabilities. The simulator provides a realistic virtual framework where users can interact with the underwater world in a 3-D and immersive setting, improving environmental awareness and dexterity, as shown in Fig. \ref{fig: wreck}.

\begin{figure}[thpb]
    \centering
    \includegraphics[width=3.4in]{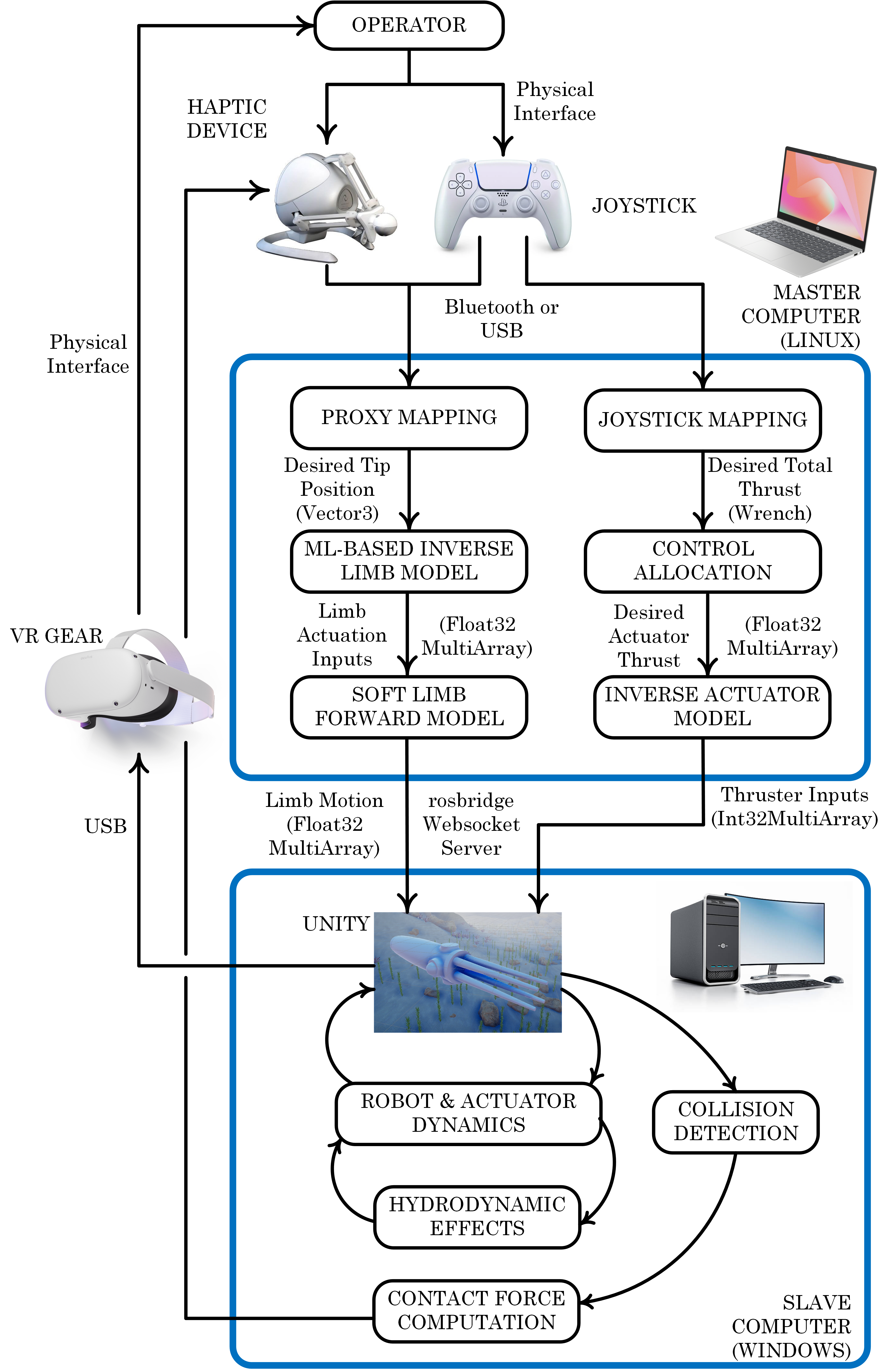} 
    \caption{A schematic of the basic building blocks of the (MARUN)$^2$ simulation environment, highlighting the Unity and ROS modules as well as VR and haptic feedback capabilities.}
    \label{fig: flowChart}
\end{figure}

(MARUN)$^2$ is designed to integrate seamlessly into existing URSULA software and UI. To this end, (MARUN)$^2$'s Unity module represents a one-to-one replacement of URSULA hardware. Guidance, navigation, command, communication, and low- and high-level control algorithms and protocols are implemented in ROS. The web-based UI developed to interact with URSULA is extended to support the (MARUN)$^2$ simulator. Finally, (MARUN)$^2$ supports VR gear for an immersive operator experience. A concise representation of the main building blocks and functionality of (MARUN)$^2$ is presented in Fig. \ref{fig: flowChart}. In what follows, these aspects of URSULA pertinent to (MARUN)$^2$ are described in detail.  

\subsection{Unity}


The Unity game engine serves as the core visualization and simulation framework in (MARUN)$^2$. Unity's extensive built-in modeling and physics-based simulation capabilities enable relatively straightforward development and modification of sophisticated and realistic dynamic virtual marine environments and worlds. 



In line with the approach presented in \cite{Haz24}, each of the four soft robotic limbs in URSULA is represented as a series of connected rigid objects to simulate continuous flexible motion. The position and rotation of the virtual limb segments are continuously updated on the basis of data received from ROS via a WebSocket connection. In addition, an orientation offset is used to align Unity’s coordinate system with ROS. A Python script processes ROS messages and maps the received coordinates to each segment, ensuring that the motion of the virtual limbs in Unity matches the received commands from the operator.

The built-in functionality of Unity is used for collision handling. Collisions and sustained rigid body contact are detected and resolved through rigid-body physics components and mesh colliders. Custom C\# scripts are utilized to compute contact forces. Unity functions such as OnCollisionEnter, OnCollisionStay, and OnCollisionExit are used to compute initial impact forces using impulse-based physics, to track continuous force application while objects remain in contact, and to reset the contact force tracking when the contact ends, respectively. This approach enables interaction with underwater objects and obstacles, realistic computation of collision impulses and contact force distribution, as well as reactive adjustments to responsive objects upon external impact.

Robot dynamics, including hydrodynamic effects (buoyancy, added mass, and hydrodynamic damping), and actuator dynamics are incorporated into Unity through custom C\# scripts. To simulate the gradual acceleration of the robot, (MARUN)$^2$ implements a progressive speed ramping mechanism. 

The current version of (MARUN)$^2$ does not incorporate any sensor models. However, models for common navigation sensors such as inertial measurement units (IMU), depth sensors, global navigation satellite system (GNSS) sensors, as well as more sophisticated acoustic and optic devices such as doppler velocimeter log (DVL), forward looking sonar (FLS), 3-D cameras, force/pressure and tactile sensors will be gradually added to the software.



\subsection{Robot Operating System}


The mission software of (MARUN)$^2$ is built on ROS Noetic, handling mission critical tasks such as waypoint-based and time-dependent trajectory-based guidance, autonomous or manual navigation, low and high-level motion control, sensor integration and fusion, power distribution and battery management, situational awareness, and surface communication. Furthermore, the mission software also enables the synchronous and coordinated teleoperation of URSULA's four soft robotic limbs when the robot is operating in an intervention mode. Publisher-subscriber pairs are utilized for bidirectional communications. The (MARUN)$^2$ is designed to operate together with URSULA's ROS-based mission software. 

URSULA follows a multimaster, multislave, bilateral teleoperation architecture with haptic feedback support. Each of URSULA's four soft robotic limbs is controlled either using a joypad (no haptic feedback) or a haptic device. Commands received from these physical interfaces are mapped to desired limb shapes in the `Proxy Mapping' node (see Fig. \ref{fig: flowChart}). Each of the four soft robotic limbs utilizes a four cable-based tendon-driven actuation system, requiring precise tendon force or displacement inputs to achieve the desired limb shapes. Due to the high complexity of the dynamic model of the limb, machine learning methods are employed in inverse kinematic and dynamic models used to compute the necessary tendon forces for a target shape. These models are indicated as `ML-Based Inverse Limb Model' node in the schematic provided in Fig. \ref{fig: flowChart}. Finally, the actual shape of the limbs are computed using a forward kinematic/dynamic model (referred to as `Soft Limb Forward Model' node) within ROS before the limb is discretized and actual positions of the limb sections are sent to Unity. 




The WebSocket-based ros\_bridge library enables continuous data transfer and synchronization between ROS software and the (MARUN)$^2$ simulation environment. This WebSocket communication protocol ensures low-latency updates for robot motion and limb control feedback loops and a straightforward integration with the (MARUN)$^2$ simulator.


\subsection{User Interface}
A web-based UI is developed for commanding and controlling URSULA. The software represents an advanced, multifunctional, and distributed UI for the commanding and controlling the robot. A web-based interface was chosen to provide mobility and ease of access from various client platforms. Although designed primarily for project URSULA, the design, layout, and functionality of the UI is kept broad so that the same interface can be utilized for command, control, and communication with a heterogeneous fleet of dissimilar vehicles including surface and subsurface marine vessels, as well as aerial vehicles such as drones.

The UI software consists of a wet-end server and a dry-end client. On the wet-end, a Node.js-based web-server runs on the embedded mission computer of the robot. The rosbridge message protocol is used to exchange information and data between on-board ROS nodes and the web server in JSON format. These JSON messages are converted to ROS topics and services using the rosbridge\_library package. A separate package called rosbridge\_server is used to send and receive JSON based messages over the WebSocket communication protocol. The roslibjs javascript library is used for the Node.js web server to correctly interpret the rosbridge messages. All the above-mentioned server-side communication software is bundled in a ROS meta-package called rosbridge\_suite. \cite{rosbridge} For low latency, bilateral, real-time communication between the wet-end server and the dry end client, the Socket.IO \cite{socketIO} library is utilized. The PM2 process management software \cite{pm2} is incorporated into the wet end web server for continuous and uninterrupted operation, as well as process monitoring and logging.

\begin{figure}[thpb]
    \centering
    \includegraphics[width=3.4in]{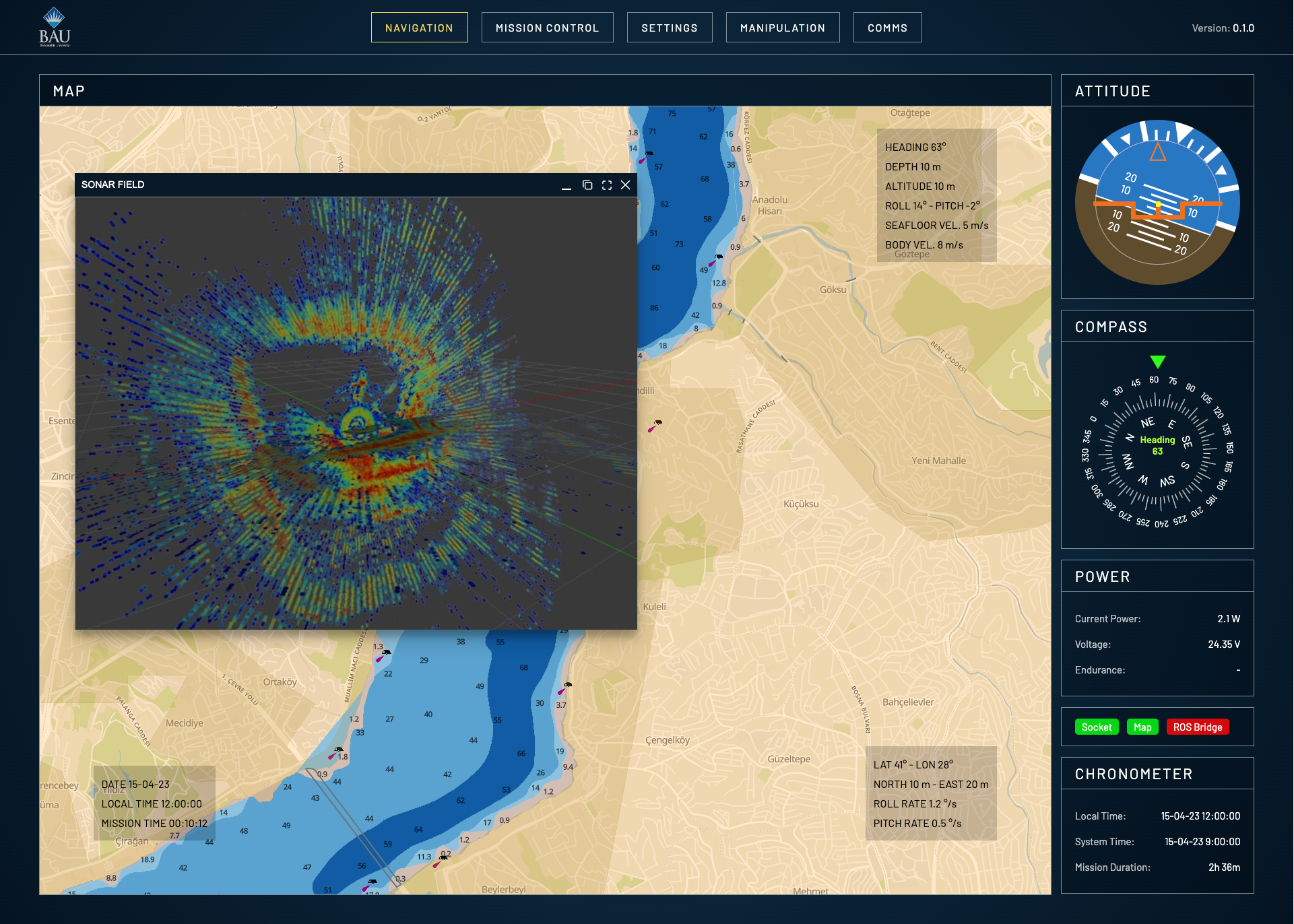} 
    \caption{An early version of the navigation module of the web-based UI developed for URSULA. In the simulation module, the map of the navigation module is replaced with a Unity render of the robot and its environment.}
    \label{fig: webUI}
\end{figure}

The dry-end, client side is centered around a dynamic and multifunctional webpage-based UI. This user interface includes several functionally distinct modules such as the Navigational Module, the Control and Situational Awareness Module, the Intervention Mode Control Module, the Mission Planning Module, and the Simulation Module (see Fig. \ref{fig: webUI}). Communication with the wet-end web-server is accomplished using the Socket.IO client library. Vue.js \cite{vueJS} is used to develop interactive tools for the webpage-based UI.        

\subsection{Virtual Reality}

The (MARUN)$^2$ simulation framework integrates VR to enhance operator environmental awareness and control efficiency in underwater robotic teleoperation. Traditional 2-D interfaces are often limited by restricted depth perception and environment awareness, which can hinder precise and dexterous manipulation in complex underwater environments. (MARUN)$^2$ uses VR to address these challenges, offering interactive visualization capabilities.

(MARUN)$^2$ employs high-resolution VR headsets that provide low latency performance. The headset is connected to the simulator computer via a USB Type-C connection, allowing high-bandwidth data transfer between the VR interface and the Unity simulation environment. The details of the VR module of (MARUN)$^2$ are depicted in Fig. \ref{fig: flowChart}.


\section{RESULTS AND DISCUSSIONS} \label{Sec: results}

The validation of (MARUN)$^2$ as a VR-enhanced underwater robotic simulation environment was carried out through controlled experiments that closely mirrored real-world underwater manipulation tasks. These experiments focused on underwater navigation, limb control, task execution efficiency, and environmental interactions. A preliminary user study with 10 subjects was conducted to compare the VR interface with traditional 2-D screen interfaces for performing dexterous underwater manipulation and seabed intervention tasks. \cite{El24} 

A series of controlled simulation tests were performed in the Unity environment to assess the movements of the limbs, the interaction with objects, and the responsiveness. The primary validation approaches included:

\begin{itemize}
    \item Trajectory-based performance evaluation: The robot's limbs are commanded to interact with test objects, while Unity recorded positional and temporal data. 
    \item Collision detection and response assessment: The impact of robotic limbs on several underwater objects was simulated.
    \item Force feedback estimation: Contact forces were approximated using rigid-body physics and impulse calculations.
\end{itemize}

\begin{figure}[thpb]
    \centering
    \includegraphics[width=3.4in]{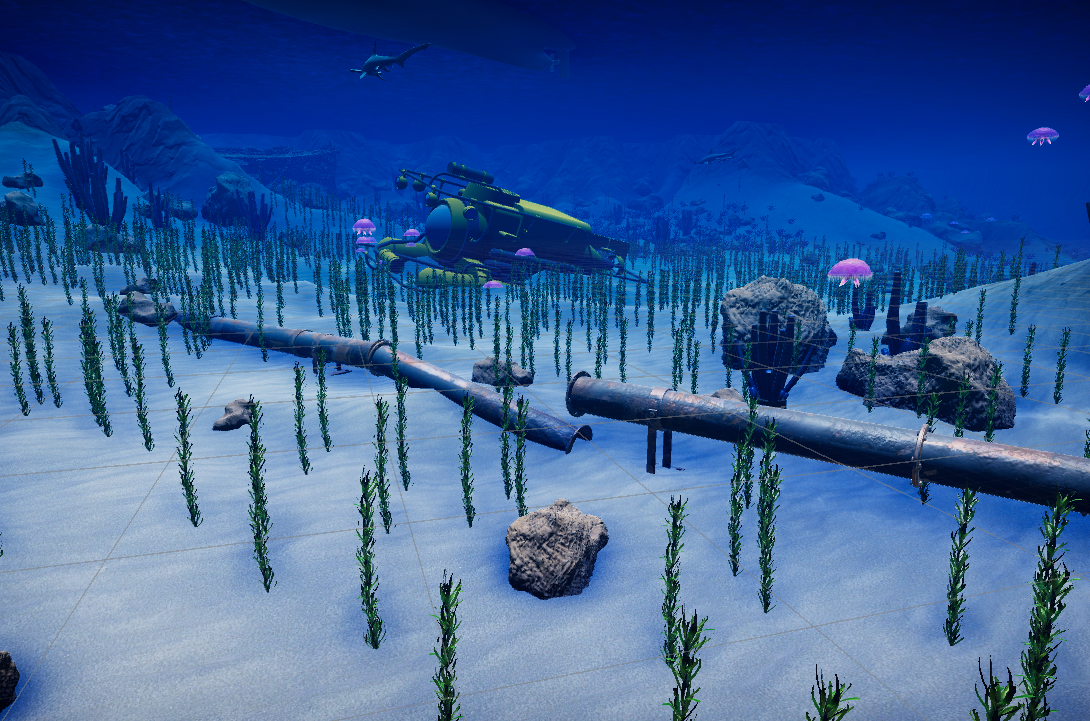} 
    \caption{A typical view of the seabed in the (MARUN)$^2$ underwater simulator, depicting a pipeline structure as well as various interactive objects.}
    \label{fig: environment2}
\end{figure}

The simulation results confirmed that (MARUN)$^2$ accurately replicates tendon-driven soft-limb movements. Fig. \ref{fig: environment2} illustrates the diverse simulated underwater environment within (MARUN)$^2$, including pipelines and other interactive man-made or natural structures and objects to evaluate the effectiveness of robotic interventions in realistic scenarios.

Three distinct test scenarios were designed to evaluate the system's precision, object manipulation, and adaptability. These tests are explained in detail below.

\begin{itemize}
    \item Test 1 -- Object Contact Task: The participants controlled the robotic limbs to make contact with a moving object (sphere), as illustrated in Fig. \ref{fig: task1}. The displacement of the sphere upon impact was analyzed to verify the precision of the interaction.

    \begin{figure}[thpb]
    \centering
    \includegraphics[width=3.4in]{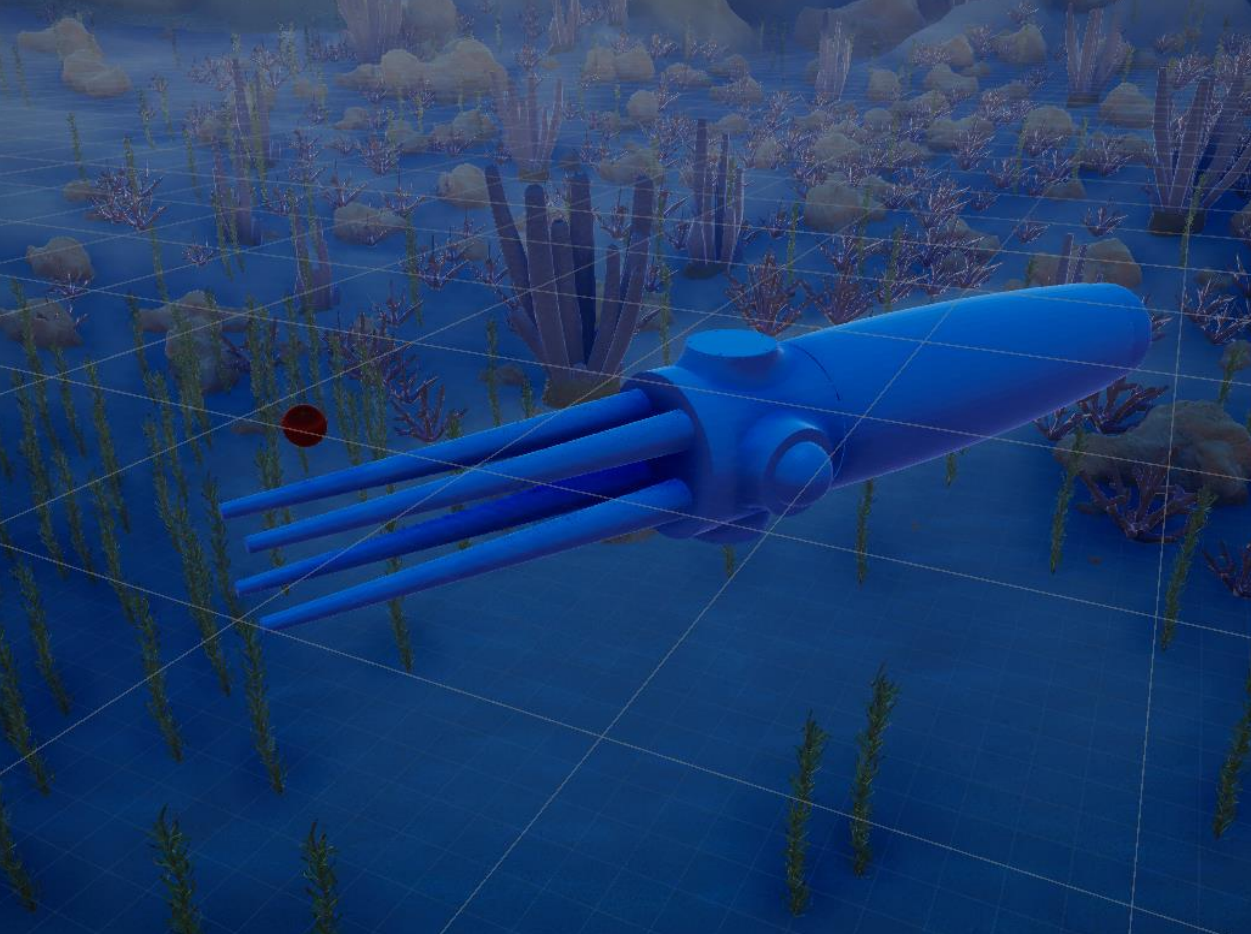} 
    \caption{Visualization of URSULA robotic limbs reaching out to a moving object (sphere) during Test 1.}
    \label{fig: task1}
    \end{figure}

    \item Test 2 -- Object Grasping Task: The robot's limbs were used to pick up objects and place them in a predefined area, as demonstrated in Fig. \ref{fig: task2}. Success rates and time-to-completion were recorded.

    \begin{figure}[thpb]
    \centering
    \includegraphics[width=3.4in]{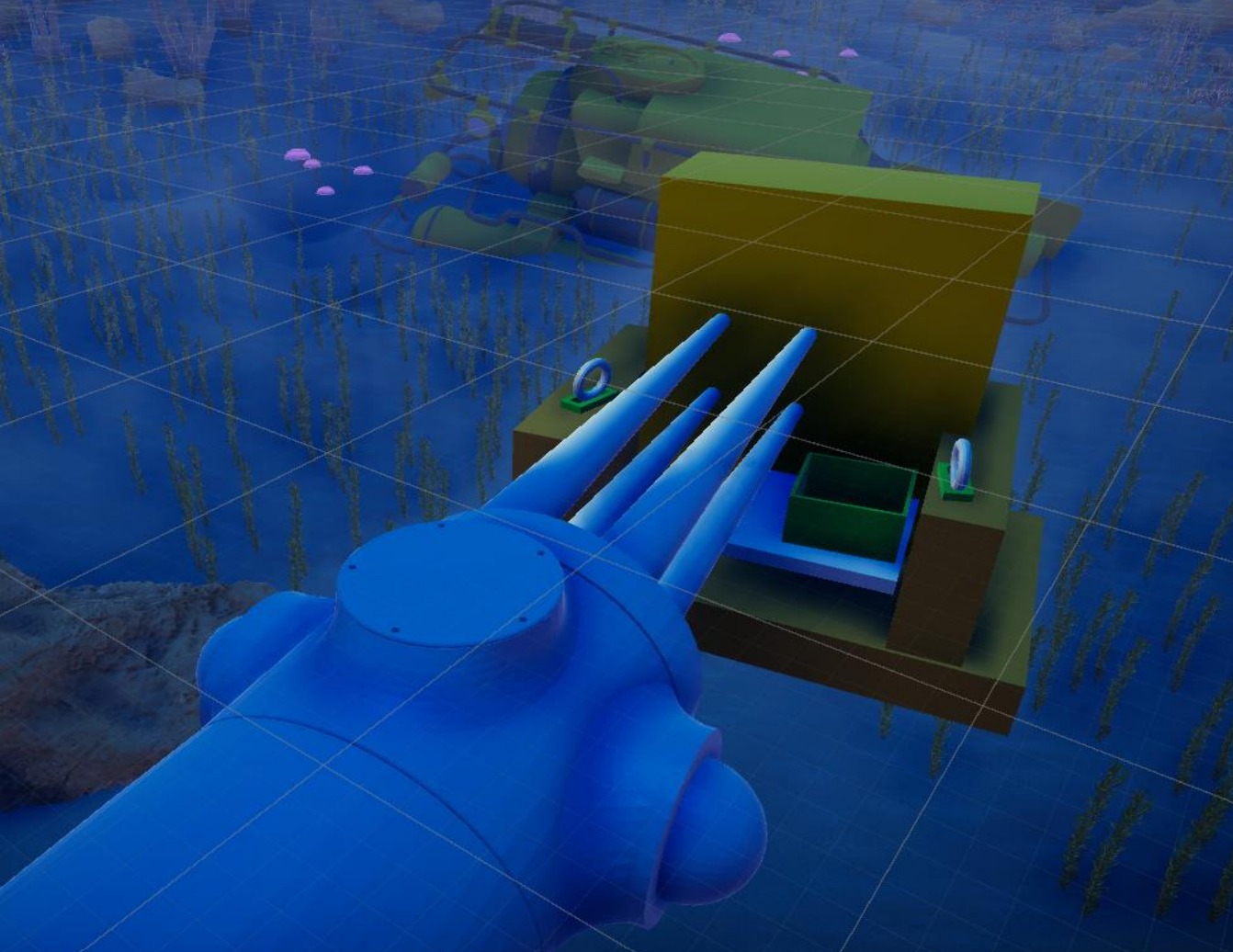} 
    \caption{View of the robotic limbs interacting with an object during Test 2.}
    \label{fig: task2}
    \end{figure}

    \item Test 3 -- Dynamic Underwater Task: The robot performed interventions in flowing water conditions, demonstrating stability and adaptive control capabilities.

\end{itemize}

Quantitative analysis of the simulation data was conducted using MATLAB. The primary performance metrics included:
\begin{itemize}
    \item Path-Length Analysis: The length of the trajectory traversed by the robotic limbs was compared across different scenarios to assess movement efficiency.

    \item Time-to-Completion: Task completion durations were recorded and analyzed.
\end{itemize}

Furthermore, qualitative data was collected from participants after the experiments were completed in the form of a 29-question survey. Each question of the survey was carefully designed and formulated to measure specific goals to gather comparison data between the normal (2-D screen) and VR modes of operation. In addition to collecting demographic information, the survey included questions aimed at measuring realism, mental/sensory effort, task difficulty, ease of use and control, level of engagement, task completion and performance satisfaction, dizziness, and discomfort level, as well as preference over the two interfaces. 

The survey results indicate that users found that the VR mode presented a more realistic telepresence experience (45\% improvement) and a higher level of engagement (37.5\% improvement), as well as easier use and control (47.5\% improvement). Mental/sensory effort (17.5\% improvement) and perceived task difficulties (42.5\% to 50\% improvement) also improved with VR. Overall, users strongly leaned towards the VR interface, which received a preference score of 4.5 where scores of 1 and 5 indicated a complete preference for the 2-D screen and VR interfaces, respectively. Task completion times were 26.75\%, 23.39\%, and 36.86\% shorter when the VR headset was used, compared to the normal mode. A more detailed presentation and analysis of the results will be provided in a future publication. 

In general, these results strongly indicate the effectiveness of VR interfaces in underwater manipulation activities over conventional 2-D screens. However, a clear variability in participant performance was also observed. This variability underscores the importance of individual skills and differences in teleoperation tasks, suggesting that while VR can enhance certain aspects of the teleoperation experience, it does not do so uniformly for all users or tasks. Personal proficiency, adaptability to VR, and the specific nature of the task all appear to influence performance outcomes. The data supports the notion that VR technology can enhance certain aspects of teleoperation while highlighting the need for a deeper understanding of how it affects manipulation and intervention strategies.

\section{CONCLUSIONS} \label{Sec: conclusions}

This paper presents the (MARUN)$^2$ simulation framework, developed to support the URSULA underwater robotic system in performing dexterous underwater manipulation and seabed intervention tasks. (MARUN)$^2$ enables the teleoperation of soft robotic limbs within a physics-accurate simulation environment. The system successfully replicates tendon-driven actuation of soft limbs, collision interactions, and task-specific manipulation scenarios. The relation between URSULA and (MARUN)$^2$ is fundamental, as the simulator serves as a testing and validation platform to refine command and control strategies before real-world deployment. Furthermore, the results indicate that VR enhances operator efficiency, spatial awareness, and engagement. However, some users reported increased dizziness, which highlighted the need for further ergonomic refinements. (MARUN)$^2$ demonstrates how VR-based teleoperation can significantly improve dexterous underwater robotic manipulation, making it a promising alternative to traditional physical UI.

Future improvements will focus on the integration of various sensors, haptic feedback capabilities, enhanced hydrodynamic modeling, and expanded actuation modes and models to further improve realism and adaptability. These advancements will solidify (MARUN)$^2$ as a comprehensive underwater teleoperation research platform.

\addtolength{\textheight}{-12cm}   


\end{document}